\documentclass[sigconf]{acmart}

\AtBeginDocument{%
  \providecommand\BibTeX{{%
    \normalfont B\kern-0.5em{\scshape i\kern-0.25em b}\kern-0.8em\TeX}}}

\usepackage{algorithm}
\usepackage{algorithmic}
\begin{document}

\title{Modeling Pedestrian Intrinsic Uncertainty for Multimodal Stochastic Trajectory Prediction via Energy Plan Denoising}

\author{Yao Liu}
\affiliation{%
  \institution{School of Computing, Macquarie University}
  \city{Sydney}
  \country{Australia}}
\email{y.liu@mq.edu.au}

\author{Quan Z. Sheng}
\affiliation{%
  \institution{School of Computing, Macquarie University}
  \city{Sydney}
  \country{Australia}}
\email{michael.sheng@mq.edu.au}

\author{Lina Yao}
\affiliation{%
  \institution{Data 61, CSIRO \& School of Computer Science and Engineering, University of New South Wales}
  \city{Sydney}
  \country{Australia}}
\email{lina.yao@unsw.edu.au}
\renewcommand{\shortauthors}{Yao Liu, et al.}


\begin{abstract}
Pedestrian trajectory prediction plays a pivotal role in the realms of autonomous driving and smart cities. 
Despite extensive prior research employing sequence and generative models, the unpredictable nature of pedestrians, influenced by their social interactions and individual preferences, presents challenges marked by uncertainty and multimodality.
In response, we propose the Energy Plan Denoising (EPD) model for stochastic trajectory prediction. 
EPD initially provides a coarse estimation of the distribution of future trajectories, termed the Plan, utilizing the Langevin Energy Model. Subsequently, it refines this estimation through denoising via the Probabilistic Diffusion Model. 
By initiating denoising with the Plan, EPD effectively reduces the need for iterative steps, thereby enhancing efficiency.
Furthermore, EPD differs from conventional approaches by modeling the distribution of trajectories instead of individual trajectories. 
This allows for the explicit modeling of pedestrian intrinsic uncertainties and eliminates the need for multiple denoising operations. 
A single denoising operation produces a distribution from which multiple samples can be drawn, significantly enhancing efficiency.
Moreover, EPD's fine-tuning of the Plan contributes to improved model performance.
We validate EPD on two publicly available datasets, where it achieves state-of-the-art results. Additionally, ablation experiments underscore the contributions of individual modules, affirming the efficacy of the proposed approach.
\end{abstract}

\ccsdesc[500]{Computing methodologies~Computer vision tasks}



\keywords{Trajectory Prediction, Diffusion Models, Energy-based Models}



\maketitle

\section{Introduction}

Trajectory prediction plays a pivotal role in various systems such as autonomous driving and smart cities, where humans are the predominant agents in their environment~\cite{survey1, TrajNet, mm-2-crime, mm-5-tracking}.
Despite significant advancements in recent years, accurately forecasting the future trajectories of pedestrians remains challenging due to the inherent multimodality and uncertainty associated with human movement. 
Humans exhibit complex behaviors such as following peers or avoiding obstacles, which can influence their trajectories. 
Moreover, individuals can spontaneously alter their trajectories based on their intentions. Consequently, owing to the stochastic nature of human behavior, even trajectories with identical past histories may diverge in their future paths~\cite{yao}.

The volition behind human movement is intrinsic, making it difficult to infer historical trajectories solely from external observations for prediction purposes. 
Consequently, in the realm of stochastic trajectory prediction, it has become customary to adopt a multimodal approach. This entails predicting the distribution of potential future trajectories rather than singular trajectories themselves. 
This approach is grounded on the premise that humans may adhere to one of several plausible trajectories encapsulated within the distribution~\cite{DESIRE, Social-GAN}.

\begin{figure}[htbp]
    \centering
    \includegraphics[width=\linewidth]{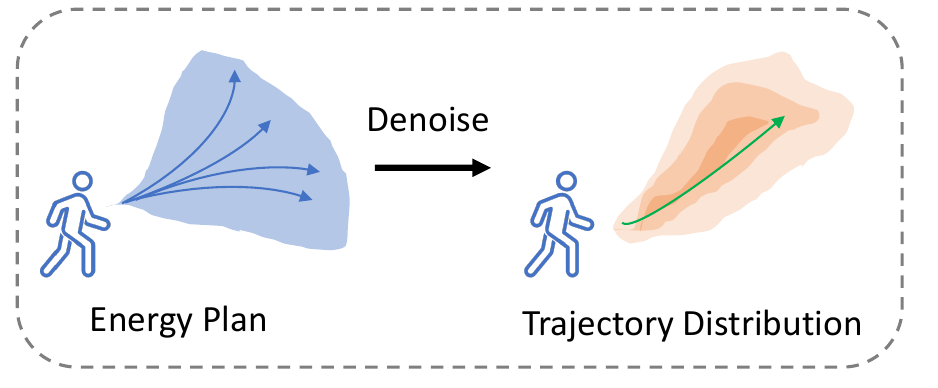}
    \caption{
    Our model initially estimates the distribution of trajectories coarsely using the Energy Model and then predicts future trajectories through denoising and sampling.
    } 
    \label{fig_intro}
\end{figure}


Previous approaches to trajectory prediction employing generative models include Generative Adversarial Networks (GANs)\cite{MG-GAN, Social-GAN, SoPhie} and Conditional Variational Autoencoders (CVAEs)\cite{DESIRE, Trajectron, Trajectron++}. 
Despite their notable achievements, these methods still encounter certain limitations: GANs' training process can be prone to instability, while CVAEs often generate trajectories that appear unnatural~\cite{mid}.

The latest diffusion models, inspired by non-equilibrium thermodynamics, have garnered significant attention in the domain of generative modeling. 
Innovations like MID~\cite{mid} and Leapfrog~\cite{Leapfrog} have introduced diffusion models to trajectory prediction, offering more efficient training compared to GANs and the ability to generate higher-quality samples than CVAEs. 
Nonetheless, diffusion models come with inherent drawbacks in the context of trajectory prediction. 
Firstly, the denoising process is time-consuming, typically requiring approximately 100 iterations to recover a clear trajectory from Gaussian noise~\cite{mid}. Secondly, unlike image generation, generating a limited number of independently distributed trajectory samples may not afford the generative model with sufficient modality~\cite{Leapfrog}.

In addition to predicting trajectories from scratch, modeling task-dependent goals is considered advantageous for capturing uncertainties and enhancing training convergence.
Several studies~\cite{PRECOG} have focused on goal-conditioned approaches, where the endpoint or certain nodes of the trajectory are coarsely estimated first, followed by predicting the entire trajectory.

This approach, known as the endpoint-based approach or goal-conditioned prediction, has been explored by PECNet~\cite{endpoint}, which generates the endpoint using CVAEs, while the subsequent LB-EBM~\cite{LB-EBM} generates the plan using an energy-based model. 
Energy models map latent vectors to their probabilities in a manner that is not constrained to be highly expressive, enabling them to capture the multimodality of trajectory distributions effectively.

To tackle the challenges posed by trajectory multimodality and uncertainty, we propose the Energy Plan Denoising (EPD) model, which focuses on modeling the distribution of trajectories rather than individual trajectories themselves, as depicted in Fig.~\ref{fig_intro}. 
Our model aims to predict the distribution of future trajectories, wherein sampling from this distribution not only incorporates inherent human uncertainty but also mitigates the computational resource overhead associated with multiple denoising processes.


Overall, we employ Diffusion to model the distribution features of pedestrian trajectories instead of modeling the trajectories directly. 
This contrasts sharply with methods like MID~\cite{mid}, which apply Diffusion models directly for trajectory prediction. 
This approach benefits from allowing the model to capture the inherent uncertainty of pedestrian movements, and using distribution sampling substantially reduces the need for inverse diffusion and denoising, thereby enhancing prediction speed. 
Our model optimizes efficiency by requiring only a single denoising operation per trajectory in multimodal prediction scenarios, a marked improvement over methods like MID, where computational demand scales with the number of trajectories.

Furthermore, our method initiates denoising not from Gaussian noise but from a trajectory estimate generated by a specially designed Energy Model. 
This significantly cuts down on the iterative denoising steps needed, unlike starting from random Gaussian noise. 
The use of an estimated trajectory as the starting point streamlines the process and boosts efficiency.

Our Energy Model incorporates strategies such as Experience Replay, which yields higher quality initial points for denoising compared to those from models like Leapfrog~\cite{Leapfrog}. 
This advantage stems from the Energy-Based Model's (EBM) differentiation from traditional Inverse Reinforcement Learning (IRL) or Generative Adversarial Imitation Learning (GAIL). 
Traditional IRL is computationally intensive as it learns cost functions in the outer loop and derives policies in the inner loop. 
In contrast, GAIL's multimodal modeling is implicit and depends on a policy generator. 
However, EBM operates in lower dimensions, is simpler to learn than IRL, and explicitly handles multimodal modeling.

Our contributions are as follows:
\begin{itemize}
    \item We introduce the Energy Plan Denoising (EPD) model for stochastic trajectory prediction. Unlike approaches that predict exact trajectories, our model predicts the distribution of future trajectories, thereby accommodating human uncertainty alongside the multimodality of trajectories.
    \item The Trajectory Distribution Model captures future trajectory features, effectively guiding the Langevin Energy Model to encompass sufficient modalities within a limited number of independently distributed trajectories. Subsequently, the high-quality coarse trajectories generated by the Langevin Energy Model, facilitated by experience replay, significantly expedite the denoising process of the Probabilistic Diffusion Model.
    \item Our model significantly enhances prediction efficiency. Validation across multiple datasets demonstrates that our model achieves state-of-the-art performance levels. Moreover, ablation experiments underscore the distinct contributions of each module.

\end{itemize}

\section{Related Works}

\subsection{Trajectory Prediction}

Trajectory prediction involves observing historical trajectories and forecasting future trajectories.

Early methods such as Social-Force~\cite{Social-force} relied on modeling attraction and repulsion forces to simulate human social interactions for predictions. 
Social-LSTM~\cite{Social-LSTM} introduced deep learning into trajectory prediction by incorporating social pooling mechanisms to aggregate interaction information for prediction. 
Following this, subsequent models have leveraged various techniques for aggregating social interaction data, including attention mechanisms~\cite{Social-Attention, mm-3-traj} and spatio-temporal graph models~\cite{Social-STGCNN}.

More recently, Transformer architectures have been adopted for trajectory prediction tasks~\cite{Transformer1}, capitalizing on their self-attention capabilities for robust temporal modeling and their suitability for parallel processing. This has allowed for more sophisticated handling of complex time dependencies in trajectory data.

In addition, some approaches formulate trajectory prediction as a generative task. 
For instance, Social-GAN~\cite{Social-GAN} proposes a pooling process to synthesize neighbor information and utilizes Generative Adversarial Networks (GANs) to predict trajectories through Recursive Neural Networks (RNNs) hidden states of neighbors.
SoPhie~\cite{SoPhie} incorporates physical and social attention modules into Long Short-Term Memory Networks (LSTMs) and employs GANs for predictions. 
AEE-GAN~\cite{mm-1-traj}, on the other hand, introduces Heterogeneous Environment information to generate future trajectories via GANs.
DESIRE~\cite{DESIRE} introduced a multimodal sampling strategy using a Conditional Variational Autoencoder (CVAE) to generate multiple future trajectories within a sampling-based Inverse Optimal Control (IOC) framework. 
Trajectron++~\cite{Trajectron++} employs Conditional Variational Autoencoders to extract potential features from past trajectories and utilizes RNNs for prediction. 
However, RNN-based methods may encounter challenges in handling sequential data with complex time dependencies due to their limited memory capacity, whereas Transformers offer enhanced modeling effectiveness and parallel computation capabilities, making them promising for trajectory prediction tasks.

\begin{figure*}[htbp]
    \centering
    \includegraphics[width=\linewidth]{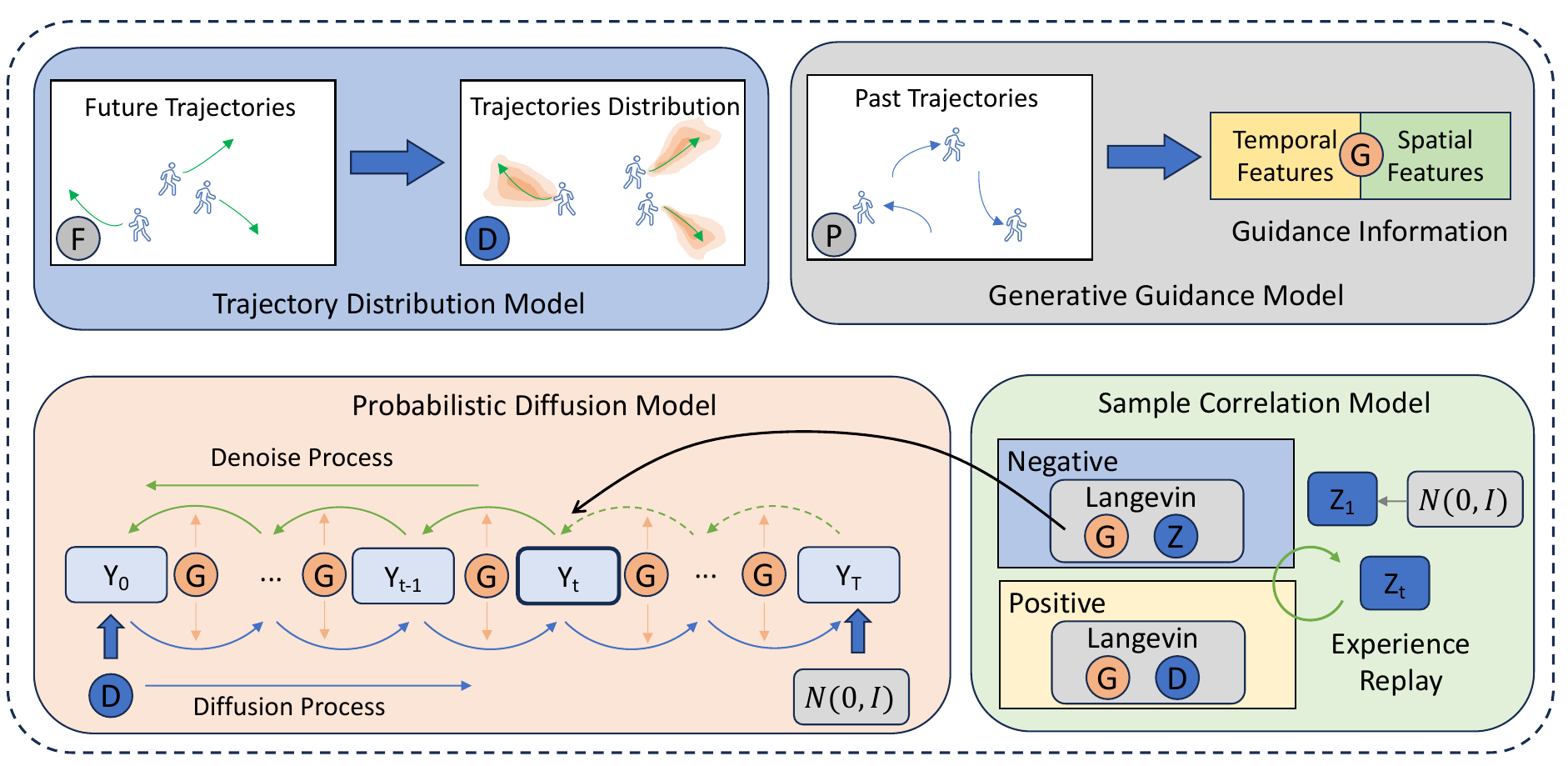}
    \caption{
    Overview of our Energy Plan Denoising (EPD) model.
    Our model comprises four main modules. The Trajectory Distribution (TD) model is utilized to model the distribution of future trajectories, while the Generative Guidance (GG) model represents the guidance information of spatio-temporal features from past trajectories. The Sample Correlation (SC) model utilizes positive and negative features alongside experience replay to generate a coarse estimate of the trajectory distribution. Subsequently, the Probabilistic Diffusion (PD) model employs the coarse estimate of the trajectory distribution as a starting point for denoising, predicting the distribution of future trajectories. It's important to note that the TD model is trained using the positive features employed for the LM model, and during inference, the SC model outputs a coarse estimate of the distribution via the negative features.
    } 
    \label{fig_overview}
\end{figure*}

\subsection{Diffusion Model}
Denoising diffusion probabilistic models (DDPM)\cite{DDPM} form the foundation of diffusion modeling, a generative approach inspired by non-equilibrium thermodynamics that has garnered considerable attention, particularly in domains such as image generation\cite{Beat-GANs}.

Some recent works~\cite{mm-4-diffusion} have introduced Diffusion models to the field of prediction, of which MID~\cite{mid} pioneered the application in trajectory prediction.
It conceptualizes trajectory prediction as a denoising process, transforming trajectories from Gaussian noise to clear paths by simulating the diffusion process—gradually adding noise until achieving a Gaussian distribution. 
MID posits that, under high uncertainty, trajectory points may be randomly distributed across the entire space. Through the denoising process, uncertainty diminishes, yielding a clear trajectory.

However, MID's efficiency is compromised due to the iterative nature of the diffusion model's denoising process, resulting in approximately 39 times the computation time of Trajectron++\cite{Trajectron++}. 
Leapfrog\cite{Leapfrog}, a subsequent model, addresses this inefficiency by initializing the denoising process with a coarsely estimated trajectory, rather than Gaussian noise. This initialization allows for a fine-tuning process, resulting in precise trajectory estimation. 
Leapfrog's denoising process is notably more efficient, requiring only a few steps, while its initializer enhances the model's ability to capture sufficient trajectory modes by incorporating trajectory mean, variance, and other features.

It's important to highlight that while Leapfrog streamlines the denoising process, it still operates on the trajectory itself. 
Moreover, despite the efficiency improvements, for multimodal trajectory prediction, multiple sampled trajectories are typically necessary for validation. 
Consequently, Leapfrog still requires multiple denoising iterations, and its computational resource consumption remains proportional to the number of samples.

\section{Methodology}

\subsection{Problem definition}

Our objective is to predict future trajectories based on past trajectory data, wherein the data solely comprises 2D coordinates of trajectory points at each time step, without additional features such as maps.

Let $x_n^t \in \mathbb{R}^2$ denote the position of the $n$-th person at time $t$. The historical trajectory of each pedestrian can be represented as $\mathbf{x}_n= \{x_n^1, \dots, x_n^{t_{\text{past}}}\}$, and the collective historical trajectory is denoted as $\mathbf{X} = \{\mathbf{x}_1, \dots, \mathbf{x}_N\}$. Correspondingly, $\mathbf{y}_n = \{y_n^{t_{\text{past}+1}}, \dots, y_n^{t_{\text{past}}+t_{\text{future}}}\}$ represents the future trajectory of pedestrians, and $\mathbf{Y} = \{\mathbf{Y}_1, \dots, \mathbf{Y}_N\}$ denotes the ensemble of future trajectories. Here, $N$ represents the number of pedestrians. Notably, since pedestrians can freely enter and exit the current space, the index $n$ may vary at different time points, implying that pedestrians may have a variable number of neighbors at each time step.

Our model predicts the distribution of future trajectories, under the assumption that pedestrians will follow one of several plausible trajectories $\hat{\mathbf{Y}}$ within this distribution.

\begin{equation}
\hat{\mathbf{Y}} \sim p(\mathbf{Y}|\mathbf{X})
\end{equation}

\subsection{Model Overview}

Our EPD model overview is depicted in Fig.\ref{fig_overview} and comprises four primary modules. Our approach goes beyond simply stacking the energy model with the diffusion model; instead, it integrates their strengths to enhance task performance. The energy model\cite{LB-EBM} is proficient in generating coarse estimated trajectories; however, subsequent refinement may not yield high-quality results. Conversely, the diffusion model is encumbered by computational demands, and while Leapfrog~\cite{Leapfrog} reduces the iterative denoising steps through initialization, it still necessitates repeated denoising for multimodal sampling.

In contrast, our EPD model leverages a coarse estimation plan as the initial denoising point for the Probabilistic Diffusion (PD) Model via the Sample Correlation (SC) Model. Additionally, it predicts the distribution of future trajectories, rather than the trajectories themselves, through the design of the Trajectory Distribution (TD) Model. This approach enables denoising the distribution of future trajectories just once for each historical trajectory, facilitating multimodal sampling efficiently.

\subsection{Trajectory Distribution Model}

Firstly, we devise a Trajectory Distribution (TD) Model to capture the trajectory features of a restricted number of independent distributions, represented by a 5-dimensional bivariate Gaussian distribution. 
For training, we directly utilize future trajectories. 
To elaborate, we initially extract the high-dimensional features of the future trajectories via the spatio-temporal graph. 
Subsequently, we encode the sufficient statistics of the distribution of these future trajectories.

\begin{equation}
p_{\alpha} =  p(\mathbf{Y}) =  Encoder_{\alpha }(ST_{\alpha }(\mathbf{Y}))
\end{equation}
We optimize $\alpha $ by minimizing the negative log-likelihood.
\begin{equation}
loss_{\alpha } = NLL(p_{\alpha}, \mathbf{Y})
\end{equation}

It's important to note that we derive the distribution of future trajectories $p_{\alpha}(\mathbf{Y})$ using the TD Model. 
This feature is solely utilized for fitting the energy model in subsequent analyses and is not employed in the inference phase.

\subsection{Generative Guidance Model}

Before constructing the Sample Correlation (SC) Model and Probabilistic Diffusion (PD) Model, it's necessary to model the features of the past trajectory $\mathbf{X}$ as guidance information $g_{\beta}$ to direct both models.

Firstly, we model the temporal features $f_{\text{temp}}$ of the past trajectories using an LSTM. 
Subsequently, we aggregate neighbor features using a social mask, and we derive the spatial features $f_{\text{spat}}$ of the past trajectories through an attention mechanism. 
Finally, we integrate these two sets of features to obtain the comprehensive guidance information $g_{\beta}$.

\begin{equation}
f_{temp}=LSTM(\mathbf{X})
\end{equation}
\begin{equation}
f_{spat}=Attn(Agg(LSTM(\mathbf{X}))
\end{equation}
\begin{equation}
g_{\beta} = (f_{temp},~f_{spat})
\end{equation}

The acquired guidance information $g_{\beta}$ encompasses the spatio-temporal features of past trajectories, which subsequently guide the generation process in both the SC Model and PD Model.


\subsection{Sample Correlation Model}

In the Sample Correlation (SC) Model, we implement the Langevin Sampling Function~\cite{LB-EBM} along with the Experience Replay Module.

Within the Langevin Function, the distribution of future trajectories is depicted by sampling $Z_\gamma^{0}$ and guidance information $g_\beta$ to generate the energy feature $Z_\gamma$. 
Additionally, we integrate Langevin experience replay by configuring the replay memory with a capacity of $K$. 
This not only stabilizes the training process, enhancing efficiency, but also facilitates learning the correlation between samples.

\begin{equation}
Z_\gamma^{0} \sim N (0,~I)
\end{equation}
\begin{equation}
Z_\gamma^k  = Langevin(Z_\gamma^{k-1}, g_\beta )  
\end{equation}

In our framework, the process of Langevin sampling $Z_\gamma$ with guidance information ($g$) is encoded through the negative encoder. 
Conversely, the distribution of future trajectories ($p_{\alpha}$) with guidance information ($g$) is encoded via the positive encoder.

\begin{equation}
Z^+=EBM_\gamma (p_{\alpha}, g_{\beta})
\end{equation}
\begin{equation}
Z^-=EBM_\gamma (p_{\alpha}, Z_\gamma )
\end{equation}

By minimizing the disparity between positive and negative encoding, we enable the sampling of the energy model to closely match the distribution of future trajectories. 
Additionally, to enhance the model's performance, it's imperative to ensure that the output of the SC model $Z_\gamma$ concurrently minimizes the negative log-likelihood concerning the true future trajectory ($\mathbf{Y}$).

\begin{equation}
Loss_\gamma = MSE(Z^+,~Z^-) + NLL(Z_\gamma,~\mathbf{Y})
\end{equation}

\subsection{Probabilistic Diffusion Model}

We introduce a Probabilistic Diffusion (PD) Model aimed at diffusing the future trajectory distribution $p_{\alpha}$ from its starting point until it converges to Gaussian noise $d^T$. In this process, guidance information $g_\beta$ is incorporated as a feature derived from the historical trajectory.

In the reverse denoising process, we iteratively reduce uncertainty using guidance information $g_\beta$ through Gaussian noise until the distribution of future trajectories is restored to $\hat{d}^0$.

\begin{equation}
d^0=p_{\alpha}
\label{equ_diff_1}
\end{equation}
\begin{equation}
    \begin{aligned}
         d^t & =diffuse_\delta(d^{t-1},~g_\beta) \\
             & =\sqrt{\alpha^t}(d^{t-1},~g_\beta)+\sqrt{1-\alpha^t}\epsilon^t \\
             & =\sqrt{\bar{\alpha^t}}(d^{0},~g_\beta)+\sqrt{1-\bar{\alpha^t}}\epsilon 
    \end{aligned}
    \label{equ_diff_2}
\end{equation}
\begin{equation}
    \begin{aligned}
    \hat{d}^t & =denoise_\delta(d^{t+1},~g_\beta) \\
              & =\frac{1}{\alpha^t}(\hat{d}^{t+1}+\frac{1-\alpha^t}{\sqrt{1-\bar{\alpha^t}}}\epsilon_\delta(g_\beta))+\sqrt{1-\alpha^t}\epsilon
    \end{aligned}
    \label{equ_diff_3}
\end{equation}

In Eq.~\ref{equ_diff_1} to Eq.~\ref{equ_diff_3}, $\alpha$ represents the weight parameter that decreases linearly with the diffusion step, while $\epsilon$ denotes the sampling noise following a normal distribution $N (0, I)$.

The PD model is optimized by minimizing both the network outcome $\epsilon_\delta$ and the random noise $\epsilon^t$. Its network architecture utilizes a backbone network constructed on Transformer.

\begin{equation}
Loss_\delta = MSE(\epsilon_\delta, \epsilon^t)
\end{equation}

\subsection{Probabislitic Trajectory Prediction}

After training the PD model, which can generate the future trajectory distribution $p(\mathbf{\hat{Y}})$, we implement truncation in the reverse process. 
In this process, we utilize the coarsely estimated future trajectory distribution $Z_\gamma$ generated by the SC model as the starting point for the reverse process of the PD model, thereby saving multiple iterations of the denoising step. 
Throughout the denoising process, we employ guidance information to direct the denoising process effectively. 
Finally, we fine-tune the parameters to optimize the negative log-likelihood of the denoised future trajectory distribution with respect to the actual future trajectory.

\begin{equation}
\hat{d}^T=Z_\gamma
\end{equation}
\begin{equation}
p(\mathbf{\hat{Y}}) = denoise_\delta(\hat{d}^T, g_{\beta})
\end{equation}
\begin{equation}
Loss_(\beta, \gamma, \delta ) = NLL(p(\mathbf{\hat{Y}}), \mathbf{Y})
\end{equation}

In our model, the output consists of a distribution of future trajectories, from which our final trajectories are derived through sampling. As a result, the computational resources required for multimodal trajectory sampling in our model are minimal.

This stands in contrast to MID~\cite{mid} and Leapfrog~\cite{Leapfrog}, where the number of trajectory samples is typically proportional to the number of denoising processes conducted. While Leapfrog reduces the number of computational iteration steps for a single denoising process, a denoising process is still necessary for each trajectory sample.

Finally, to evaluate our model, we employ the average displacement error (ADE) and the final displacement error (FDE).

\subsection{Algorithm}
The overall prediction process of our model is shown in Algorithm~\ref{algo}.

\begin{algorithm}[ht] 
\caption{Energy Plan Denoising}
\label{algo}
\begin{algorithmic}[1]
 \renewcommand{\algorithmicrequire}{\textbf{Input:}}
 \renewcommand{\algorithmicensure}{\textbf{Output:}}
 \REQUIRE Past Trajectories
 \ENSURE  Future Trajectories
 \\ \textit{Training Process} :
  \STATE { Distribution of Trajectories ($p$) $\leftarrow$ $TD$(Future Trajectories ($\mathbf{Y}$)) }
  \STATE { Guidance Information ($g$) $\leftarrow$ $GG$(Past trajectorie ($\mathbf{X}$)) }
  \STATE { Positive and Negative Features ($z^+,z^-$) $\leftarrow$ $SC$($g$, $p$) }
  \STATE { Diffused noise ($N \in (0,I)$) $\leftarrow$ $PD$($g$, $p$, $z$) }
  \\ \textit{Inference Process} :
  \STATE { Guidance Information ($g$) $\leftarrow$ $GG$(Past trajectorie ($\mathbf{X}$)) }
  \STATE { Negative Features ($z^-$) $\leftarrow$ $SC$($g$) }
  \STATE { Predicted Trajectory Distribution ($\hat{p}$) $\leftarrow$ $PD$($g$, $p$, $z^-$) }
  \STATE { Predicted Trajectories ($\mathbf{\hat{Y}}$) $\leftarrow$ $Sample$($\hat{p}$) }
  
 \RETURN Predicted Trajectories
 \end{algorithmic} 
 
\end{algorithm}

\section{Experiments}

\subsection{Datasets and Settings}

We evaluate our model's performance using the ETH/UCY and SDD datasets. The ETH/UCY dataset comprises five sub-datasets: ETH, HOTEL, UNIV, ZARA1, and ZARA2, with each sampled at 0.4 seconds per frame. The prediction task involves forecasting the trajectory of the next 12 frames based on observations from the previous 8 frames. Evaluation follows the leave-one-out method, where each sub-dataset serves as the test set once, while the remaining four are used for training and validation.

The SDD dataset includes 20 scenes, with the prediction task entailing forecasting 12 future frames based on 8 preceding frames.

In our model, we employ an MLP as the Encoder for the distribution model. The temporal and spatial features in the guidance model have a dimensionality of 128. The replay memory size in the energy model is set to 1,000. For the diffusion model, we utilize 100 diffusion steps, with 5 steps dedicated to denoising and fine-tuning.

\subsection{Experiments}

\begin{table*}[htbp]
\caption{Pedestrian Trajectory Prediction Results on ETH/UCY. Values denote ADE/FDE in meters; * represents a preprocessing problem in the dataloader such that performance is overestimated; ${}^\dagger$ denotes additional information such as images. The best results are indicated in bold, while the second-best results are underlined.}
\label{table_result}
\centering
\resizebox{0.7\linewidth}{!}{
\begin{tabular}{l|cccccc}
\toprule
Methods                                                & ETH        & HOTEL     & UNIV      & ZARA1     & ZARA2     & AVG       \\ \midrule
Linear                                          & 1.33/2.94  & 0.39/0.72 & 0.82/1.59 & 0.62/1.21 & 0.77/1.48 & 0.79/1.59 \\ \midrule
LSTM                                            & 1.09/2.41  & 0.86/1.91 & 0.61/1.31 & 0.41/0.88 & 0.52/1.11 & 0.70/1.52 \\ \midrule
Social-LSTM~\cite{Social-LSTM}                  & 1.09/2.35  & 0.79/1.76 & 0.67/1.40 & 0.47/1.00 & 0.56/1.17 & 0.72/1.54 \\ \midrule
Social-ATTN~\cite{Social-Attention}             & 0.39/3.74  & 0.29/2.64 & 0.33/3.92 & 0.20/0.52 & 0.30/2.13 & 0.30/2.59 \\ \midrule
Social-BiGAT~\cite{Social-BiGAT}                & 0.69/1.29  & 0.49/1.01 & 0.55/1.32 & 0.30/0.62 & 0.36/0.75 & 0.48/1.00 \\ \midrule
Social-GAN~\cite{Social-GAN}                    & 0.81/1.52  & 0.72/1.61 & 0.60/1.26 & 0.34/0.69 & 0.42/0.84 & 0.58/1.18 \\ \midrule
SoPhie${}^\dagger$~\cite{SoPhie}                & 0.70/1.43  & 0.76/1.67 & 0.54/1.24 & 0.30/0.63 & 0.38/0.78 & 0.54/1.15 \\ \midrule
Social-STGCNN~\cite{Social-STGCNN}              & 0.64/1.11  & 0.49/0.85 & 0.44/0.79 & 0.34/0.53 & 0.30/0.48 & 0.44/0.75 \\ \midrule
Transformer~\cite{Transformer1}                 & 0.61/1.12  & 0.18/0.30 & 0.35/0.65 & 0.22/0.38 & 0.17/0.32 & 0.31/0.55 \\ \midrule
STAR~\cite{STAR}                                & 0.36/0.65  & 0.17/0.36 & 0.26/0.55 & 0.22/0.46 & 0.31/0.62 & 0.26/0.53 \\ \midrule
Trajectron~\cite{Trajectron}                    & 0.59/1.14  & 0.35/0.66 & 0.54/1.13 & 0.43/0.83 & 0.43/0.85 & 0.47/0.92 \\ \midrule
Trajectron++${}^*$~\cite{Trajectron++}                & 0.39/0.83  & 0.12/0.21 & 0.20/0.44 & 0.15/0.33 & 0.11/0.25 & 0.19/0.41 \\ \midrule
PECNet\cite{endpoint}                           & 0.54/0.87  & 0.18/0.24 & 0.35/0.60 & 0.22/0.39 & 0.17/0.30 & 0.29/0.48 \\ \midrule
LB-EBM~\cite{LB-EBM}                            & 0.30/0.52  & 0.13/0.20 & 0.27/0.52 & 0.20/0.37 & 0.15/0.29 & 0.21/0.38 \\ \midrule

MID${}^*$~\cite{mid}                                  & 0.39/0.66  & 0.13/0.20 & 0.27/0.52 & 0.20/0.37 & 0.15/0.29 & 0.21/0.38 \\ \midrule
Leapfrog~\cite{mid}                             & 0.39/0.58  & 0.11/0.17 & 0.26/0.43 & 0.18/0.26 & 0.13/0.22 & 0.21/0.33 \\ \midrule \midrule

Trajectron++~[repo]  & 0.67/1.18                         & \textbf{0.18}/\textbf{0.28}   & \textbf{0.30}/\underline{0.54}    & \textbf{0.25}/\textbf{0.41}       & \textbf{0.18}/\textbf{0.32}  & 0.32/0.55 \\ \midrule
MID~[repo]           & \textbf{0.54}/\underline{0.82}    & \underline{0.20}/0.31         & \textbf{0.30}/0.57                & 0.27/0.46                         & 0.20/\underline{0.37}        & \textbf{0.30}/\underline{0.51} \\ \midrule

Ours                 & \underline{0.55}/\textbf{0.81}    & 0.21/\textbf{0.28}            & \textbf{0.30}/\textbf{0.53}       & \textbf{0.25}/\underline{0.44}    & \underline{0.19}/0.40        & \textbf{0.30}/\textbf{0.49} \\ \bottomrule

\end{tabular}
}
\end{table*}

\begin{table}[htbp]
\caption{Pedestrian Trajectory Prediction Results on SDD. Values denote ADE/FDE in meters; * represents a preprocessing problem in the dataloader such that performance is overestimated; ${}^\dagger$ denotes additional information such as images.}
\label{table_sdd}
\centering
\resizebox{0.4\linewidth}{!}{
\begin{tabular}{l|c}
\toprule
Methods                                  & SDD         \\ \midrule
Social-LSTM~\cite{Social-LSTM}                  & 31.19/56.97   \\ \midrule
Social-GAN~\cite{Social-GAN}                    & 27.23/41.44   \\ \midrule
SoPhie${}^\dagger$~\cite{SoPhie}                & 16.27/29.38   \\ \midrule
Desire~\cite{DESIRE}                            & 19.25/34.05   \\ \midrule
Trajectron++${}^*$~\cite{Trajectron++}                & 19.30/32.70   \\ \midrule

PECNet\cite{endpoint}                           & 9.96/15.88   \\ \midrule
LB-EBM~\cite{LB-EBM}                            & 8.87/15.61   \\ \midrule

MID${}^*$~\cite{mid}                            & 7.61/14.30   \\ \midrule \midrule

MID~[repo]                                      & 9.08/18.06   \\ \midrule
Ours                                            & 8.97/15.72

\\ \bottomrule
\end{tabular}
}
\end{table}

\begin{figure*}[ht]
    \centering
    \includegraphics[width=1\linewidth]{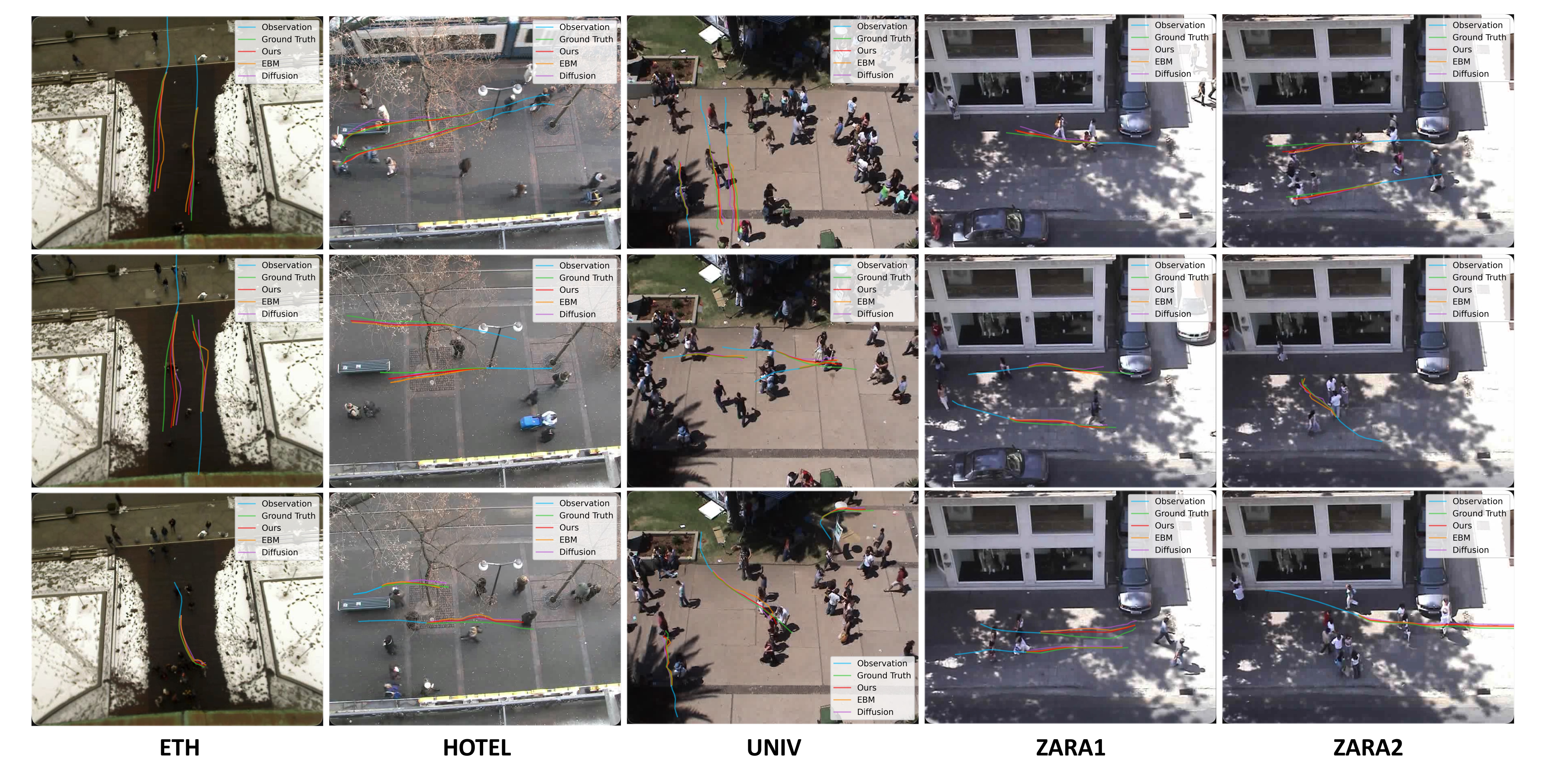}
    \caption{
    Visualization of pedestrian trajectory prediction. 
    } 
    \label{fig_comp}
\end{figure*}

\subsubsection{Comparison Experiments}

We conduct comparisons between our model predictions and those of other state-of-the-art models on both the ETH/UCY dataset and the SDD dataset.

The primary models we compare against include Trajectron++\cite{Trajectron++} and MID\cite{mid}. However, it's worth noting that Trajectron++ and MID may suffer from data preprocessing issues, leading to potential overestimation of their performance. Therefore, we compare against our reproduced results of Trajectron++ and MID.

Additionally, models such as PECNet~\cite{endpoint}, LB-EBM~\cite{LB-EBM}, and Leapfrog~\cite{Leapfrog} do not disclose their data preprocessing schemes, making fair comparison difficult.

Results for the ETH/UCY dataset are presented in Tab.~\ref{table_result}. Our reproduced results show a significant advantage over Trajectron++ and MID, particularly achieving state-of-the-art performance in terms of overall averages. Notably, our model performs exceptionally well on the UNIV subdataset, which is the most complex, as illustrated in Fig.~\ref{fig_comp}.

Results for the SDD dataset are shown in Tab.~\ref{table_sdd}, where we also outperform the reproduced MID.

\begin{itemize}
    \item Linear: This model utilizes linear regression to minimize the least square error of linear parameters.
    \item LSTM: It utilizes a standard Long Short-Term Memory Network, conducting individual regression for each trajectory without considering social interaction information.
    \item Social-LSTM~\cite{Social-LSTM}: It aggregates social interaction information through a social pooling mechanism and regresses future trajectories using LSTM.
    \item Social ATTN~\cite{Social-Attention}: It aggregates features through a social attention mechanism and regresses future trajectories using LSTM.
    \item Social-BiGAT~\cite{Social-BiGAT}: It is based on LSTM and utilizes a graph structure for prediction
    \item Social GAN~\cite{Social-GAN}: It builds upon Social LSTM, generating future trajectories using a Generative Adversarial Network (GAN).
    \item SoPhie~\cite{SoPhie}: It combines LSTM and GAN, adding physical and social attention modules to the model.
    \item Social-STGCNN~\cite{Social-STGCNN}: It relies on graph structures with Convolutional Neural Networks (CNN), omitting recurrent structures or Transformer structures.
    \item Desire~\cite{DESIRE}: It introduces an inverse optimal control-based trajectory planning method that incorporates a refinement structure for predicting trajectories.
    \item Transformer~\cite{Transformer1}: It directly employs a Transformer for the regression of future trajectories, without considering any social interaction information.
    \item STAR~\cite{STAR}: It is a prediction model that utilizes a graph structure with Transformer architecture.
    \item Trajectron~\cite{Trajectron}: It merges LSTM and variational deep generative modeling to generate the distribution of trajectories.
    \item Trajectron++~\cite{Trajectron++}: Expanding upon Trajectron, this model employs a directed-edges building block graph structure and regresses the trajectory distribution using a recurrent structure.
    \item PECNet~\cite{endpoint}: It represents a pioneering work on goal-conditional methods, initially estimating the pedestrian’s goal and subsequently predicting the trajectory conditional on the goal.
    \item LB-EBM~\cite{LB-EBM} It is a probabilistic model with a cost function defined in the latent space to incorporate movement history and social context. The low dimensionality of the latent space, combined with the high expressivity of the EBM, facilitates the model's ability to capture the multimodality of pedestrian trajectory distributions.
    \item MID~\cite{mid}: It represents a recent attempt to integrate diffusion models with trajectory prediction. MID approximates reverse diffusion processes to gradually eliminate predictive uncertainty and ultimately obtain an exact predictive trajectory for each pedestrian.
    \item Leapfrog~\cite{Leapfrog}: It reduces the diffusion chain and samples coarse trajectories and random noises to ensure multi-modality. However, diffusion is applied on exact locations and executed once for each trajectory, which does not effectively decrease the overall count of reverse diffusion processes.
\end{itemize}

\subsubsection{Visualization Analysis}

\begin{figure}[ht]
    \centering
    \includegraphics[width=\linewidth]{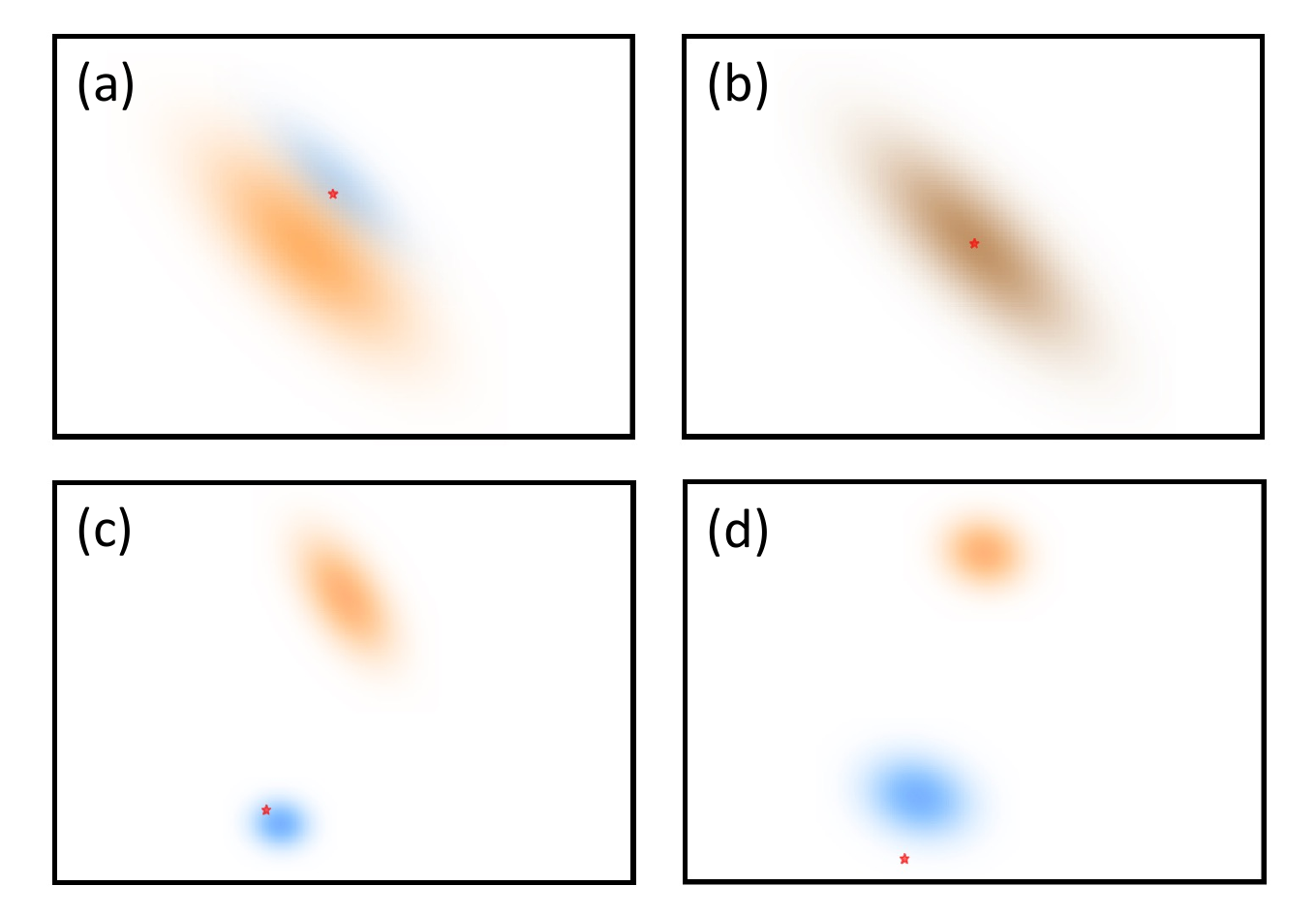}
    \caption{
    Visualization of pedestrian trajectory points with probability density distribution.  Each pedestrian location is denoted by a star symbol. The orange color corresponds to the coarsely estimated probability from the Sample Correlation (SC) Model, while the blue color indicates the probability predicted by the Probabilistic Diffusion (PD) model. Darker shades signify higher probabilities.}
    \label{fig_gau}
\end{figure}

We illustrate the probabilities estimated by the LM and PD models in Fig.~\ref{fig_gau}. In (a), we observe a scenario with minimal bias, where PD refines the LM outcomes. On the contrary, (b) illustrates a situation with virtually no bias. (c) and (d) portray instances with significant bias, where PD fine-tuning enhances the representativeness of the distributions towards the actual locations.

\begin{figure}[ht]
    \centering
    \includegraphics[width=\linewidth]{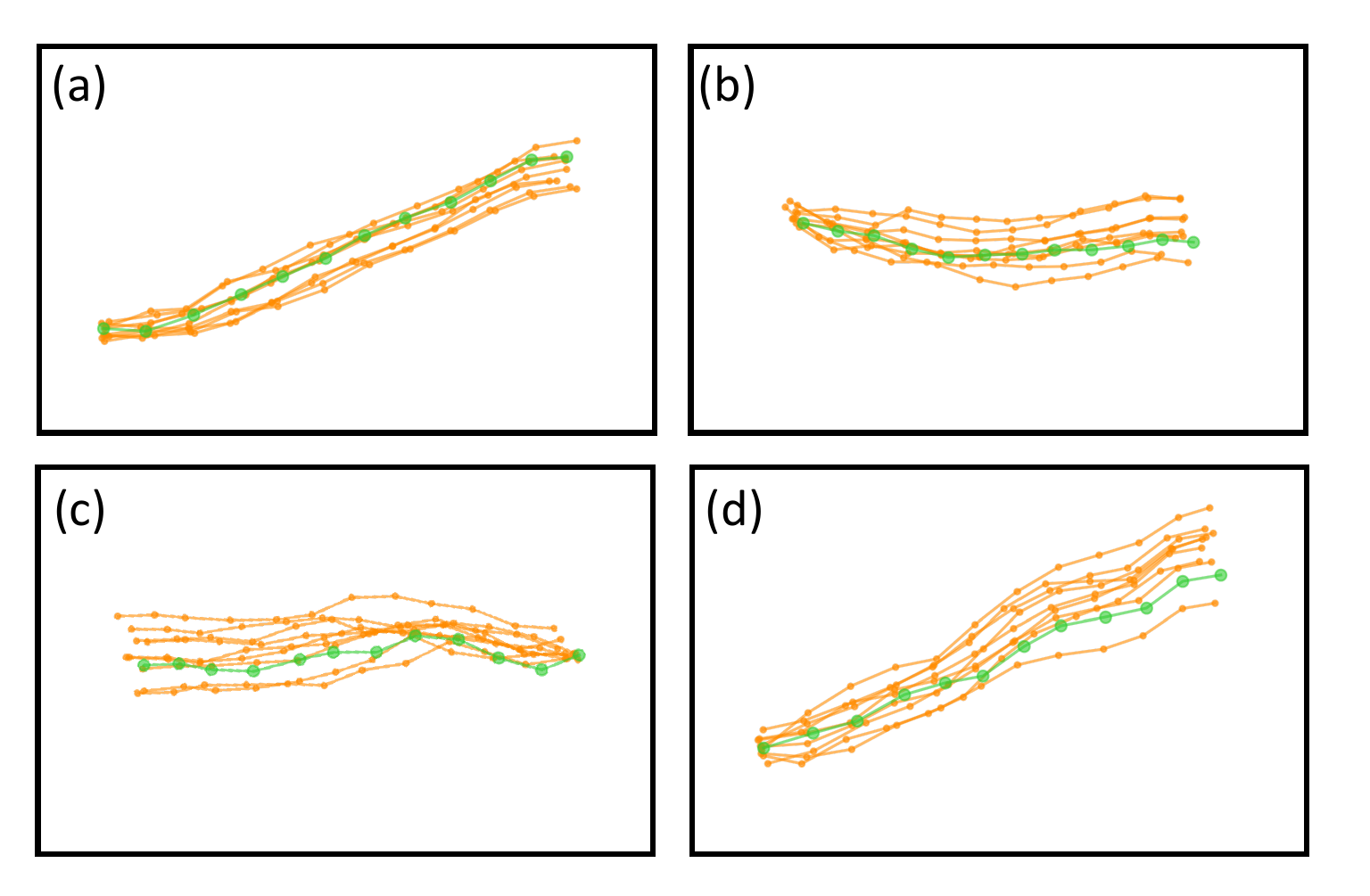}
    \caption{
    Visualization of future trajectories and sampled predicted trajectories. Green lines represent future trajectories, while orange lines denote predicted trajectories sampled from the generated distribution.
    } 
    \label{fig_traj}
\end{figure}

In Fig.~\ref{fig_traj}, we depict the sampling of the trajectory distribution alongside real trajectories. 
Notably, the sampled trajectories closely encircle the vicinity of the true trajectories, signifying the validity of the estimated trajectory distribution and its ability to encompass plausible trajectories. 
Moreover, the dispersion of trajectories over time is effectively mitigated, with trajectory points at endpoints exhibiting controlled dispersal, rather than scattering in all directions.

\begin{table}[htbp]
\caption{Model Efficiency Comparison. rel. time denotes relative time and rel. memory denotes relative memory; PD denotes Probabilistic Diffusion Model.}
\label{table_time}
\centering
\resizebox{0.5\linewidth}{!}{
\begin{tabular}{l|cc}
\toprule
Methods & rel. time     & rel. memory      \\ \midrule
EPM (w/o PD)        & 0.66 & --\\ \midrule
EPM                 & 1  & 1\\     \midrule
MID~\cite{mid}                 & 9.84 & 1.48 \\ \bottomrule
\end{tabular}
}
\end{table}

\begin{table}[htbp]
\caption{Ablation Experiments.PD denotes Probabilistic Diffusion Model; SC denotes Sample Correlation Model.}
\label{table_ablation}
\centering
\resizebox{0.7\linewidth}{!}{
\begin{tabular}{cc|ccccc}
\toprule
PD & SC & ETH       & HOTEL     & UNIV      & ZARA1     & ZARA2     \\ \midrule
w/        & w/o & 0.76/1.48 & 0.26/0.45 & 0.38/0.72 & 0.41/0.66 & 0.38/0.66 \\ \midrule
w/o       & w/  & 0.62/1.32 & 0.26/0.41 & 0.45/0.67 & 0.33/0.53 & 0.29/0.44 \\ \midrule
w/        & w/  & 0.55/0.81 & 0.21/0.28 & 0.30/0.53 & 0.25/0.44 & 0.19/0.40 \\ \bottomrule
\end{tabular}
}
\end{table}

\subsubsection{Efficiency Analysis}

We compare the efficiency of our EPD model with MID in Tab.~\ref{table_time}. Our model, initiating denoising from the coarse estimation, significantly reduces the iteration steps required for denoising and diminishes the need for repetitive denoising in overall sampling by predicting trajectory distributions. This substantial reduction in time consumption is evident. Furthermore, considering our model comprises both SC and PD modules, we conduct a comparison excluding the PD module. Remarkably, we find that the denoising facilitated by the PD module constitutes only 1/3 of the overall time.

\subsubsection{Ablation Analysis}

Lastly, we conduct ablation experiments on the EPD model. EPD without LM represents the denoising model, while EPD without PD represents the energy model. 
As depicted in Tab.~\ref{table_ablation}, both the denoising model and the energy model exhibit some capability in estimation and prediction, albeit with less satisfactory outcomes. 
Notably, our EPD model, leveraging the distribution generated by LM, significantly enhances overall performance. 
It is noteworthy that the denoising model shows certain advantages in the UNIV dataset, suggesting its adaptability to more complex tasks in certain scenarios.

\section{Conclusion}

We introduce the Energy Plan Denoising (EPD) model for stochastic trajectory prediction.
Our approach initially provides a coarse estimation of the distribution of future trajectories through energy modeling, followed by trajectory prediction via denoising and Gaussian sampling. 
By commencing with a coarse estimate, our model enhances efficiency by reducing the iterative steps of denoising. 
Additionally, our strategy of predicting trajectory distributions, rather than individual trajectories, mitigates the need for repetitive denoising during sampling. 
Through evaluation across diverse datasets, our model achieves state-of-the-art performance.

\bibliographystyle{ACM-Reference-Format}
\bibliography{main}

\end{document}